\documentclass[letterpaper]{article} 

\usepackage{aaai25}  
\usepackage{times}  
\usepackage{helvet}  
\usepackage{courier}  
\usepackage[hyphens]{url}  
\usepackage{graphicx} 
\usepackage{natbib}  
\usepackage{caption} 
\usepackage{algorithm}
\usepackage{algorithmic}
\usepackage{xcolor,colortbl}
\usepackage{amsmath}
\usepackage{amssymb}
\usepackage{listings}

\usepackage{multirow}
\usepackage{booktabs}

\usepackage{mathtools}
\usepackage{newfloat}
\usepackage{listings}
\DeclareCaptionStyle{ruled}{labelfont=normalfont,labelsep=colon,strut=off} 
\floatstyle{ruled}
\newfloat{listing}{tb}{lst}{}
\floatname{listing}{Listing}
\title{S2S2: Semantic Stacking for Robust Semantic Segmentation in Medical Imaging}
\author{
    Yimu~Pan\textsuperscript{\rm 1},
    Sitao~Zhang\textsuperscript{\rm 1},
    Alison~D.~Gernand\textsuperscript{\rm 1},
    Jeffery~A.~Goldstein\textsuperscript{\rm 2},
    James~Z.~Wang\textsuperscript{\rm 1}
}
\affiliations{
    \textsuperscript{\rm 1}The Pennsylvania State University, University Park\\
    \textsuperscript{\rm 2}Northwestern University, Chicago\\
    \{ymp5078, sitao.zhang, adg14\}@psu.edu, ja.goldstein@northwestern.edu, jwang@psu.edu
}

\usepackage{bibentry}

\begin{document}

\maketitle

\begin{abstract}
Robustness and generalizability in medical image segmentation are often hindered by scarcity and limited diversity of training data, which stands in contrast to the variability encountered during inference. While conventional strategies---such as domain-specific augmentation, specialized architectures, and tailored training procedures---can alleviate these issues, they depend on the availability and reliability of domain knowledge. When such knowledge is unavailable, misleading, or improperly applied, performance may deteriorate. In response, we introduce a novel, domain-agnostic, add-on, and data-driven strategy inspired by image stacking in image denoising. Termed ``semantic stacking,'' our method estimates a denoised semantic representation that complements the conventional segmentation loss during training. This method does not depend on domain-specific assumptions, making it broadly applicable across diverse image modalities, model architectures, and augmentation techniques. Through extensive experiments, we validate the superiority of our approach in improving segmentation performance under diverse conditions. Code is available at https://github.com/ymp5078/Semantic-Stacking.
\end{abstract}

%

\section{Introduction}
\label{sec:intro}
In the rapidly evolving field of computer vision, significant progress in image recognition has been driven by not only groundbreaking developments in model architectures~\cite{he2016deep, dosovitskiy2020image, ronneberger2015u} but also deliberated training recipes~\cite{wightman2021resnet, liu2022convnet, woo2023convnext} and innovative augmentation techniques~\cite{cubuk2020randaugment, cubuk2018autoaugment, hendrycks2019augmix, yun2019cutmix}. These advancements largely stem from the abundance and diversity of natural image datasets~\cite{russakovsky2015imagenet, lin2014microsoft, krishna2017visual}, which enable models to learn robust, generalizable features. 

In contrast, medical image analysis faces distinct challenges. Data are often scarce and originate from a limited number of sites, captured through specific imaging devices, or within certain modalities~\cite{litjens2017survey}. High annotation costs further exacerbates these challenges, making the pursuit of training robust models in medical image analysis a paramount yet elusive goal~\cite{aggarwal2021diagnostic, nguyen2023out}. The need for model robustness in medical imaging is critical: errors can have severe clinical consequences~\cite{esteva2019guide}. While augmentation techniques can mitigate data limitation, they can be inadequate, or even detrimental, if misapplied to medical contexts~\cite{perez2018data,ozbulak2019impact}. The heterogeneous nature of medical images---ranging from Computerized Tomography (CT) and Magnetic Resonance Imaging (MRI) scans to standard RGB photographs---further complicates the development of universally applicable augmentation strategies. Together, these issues underscore the urgent need for approaches that enhance model robustness without succumbing to the pitfalls of domain-specified bias~\cite{seyyed2021underdiagnosis, roberts2021common}.

\begin{figure}[!t]
    \centering
    \includegraphics[width=\linewidth]{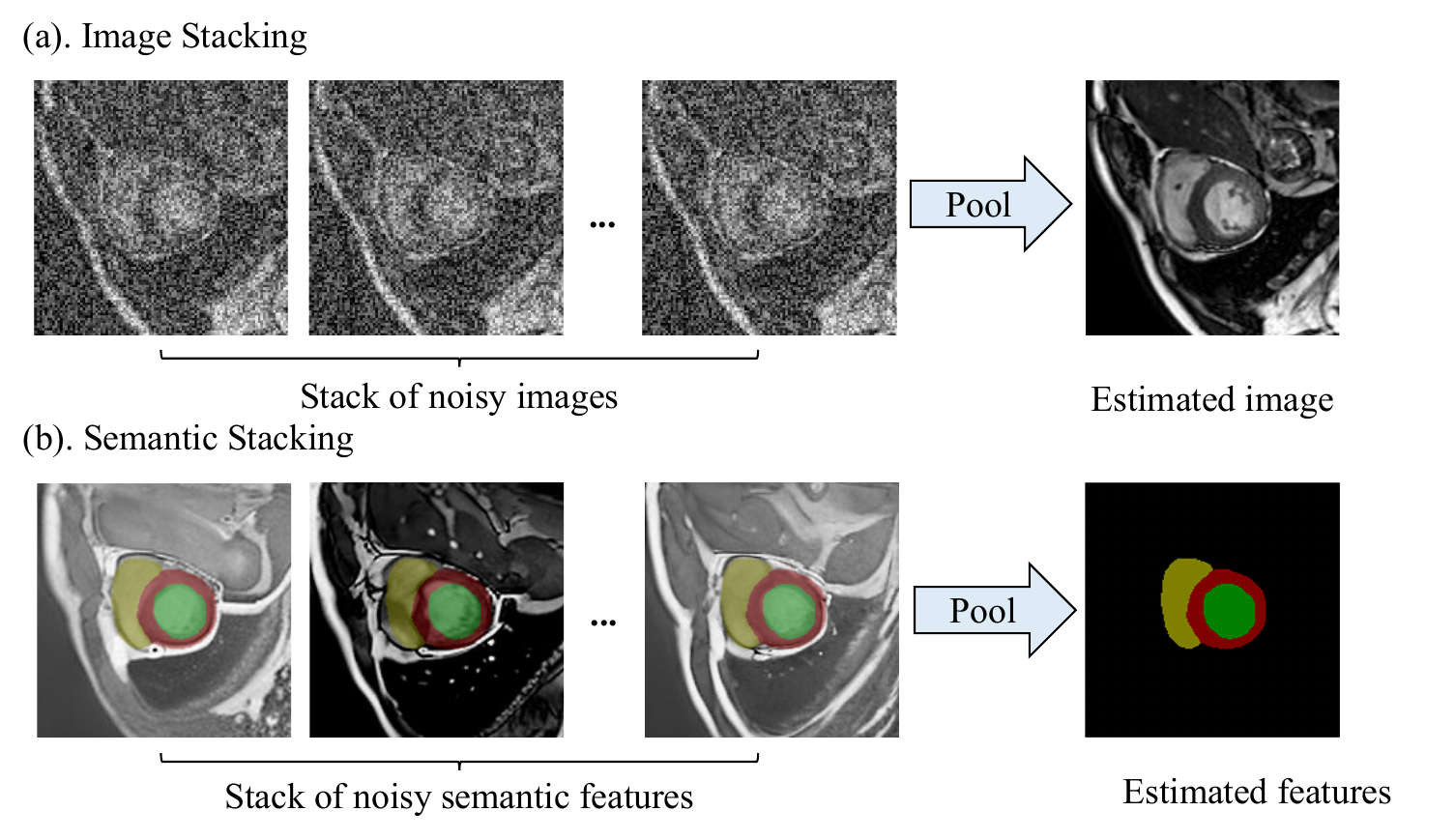}
    \caption{An illustration of the proposed semantic stacking approach compared to traditional image stacking for noise reduction. (a) Image stacking for noise reduction in imagery. (b) Our semantic stacking technique, aimed at reducing feature noise. Here, we illustrate semantic features through semantic segmentation maps for clarity, though our method operates on encoded features.}
    \label{fig:semantic_stacking_compare}
\end{figure}

In this work, we introduce an add-on training strategy, Semantic Stacking for Semantic Segmentation (S2S2), that can be seamlessly integrated into existing pipelines. Unlike previous approaches that focus narrowly on either in-domain performance~\cite{chen2021transunet} or out-of-domain robustness~\cite{su2023rethinking}, our method enhances both. Drawing inspiration from image stacking in image denoising, where multiple noisy images are stacked and pooled to estimate a denoised image, we propose {\it semantic stacking}: we first estimate a denoised semantic representation from a stack of synthetic images and then encourage the network to learn from this representation. We argue that this estimated denoised semantic representation more closely reflects the underlying ground truth, thus reducing both bias and variance. This method directs models toward a denoised semantic representation, distinguishing itself through a data-driven design that avoids domain-specific assumptions. This versatility makes our approach particularly advantageous across diverse image modalities, serving as an invaluable asset in scenarios where broad generalizability is critical and specific domain knowledge remains elusive.

Additionally, directly estimating the semantic stacking requires obtaining the semantic representation from all images in the stack. Running the network through all images in the stack at each iteration is resource- and time-intensive, or even impractical, as the stack grows. Through theoretical analysis, we derived a practical upper bound for semantic variations, transforming the semantic stacking into an operation involving only two images per iteration. This transformation makes learning from the semantic stacking feasible.

As general-purpose interactive segmentation tools gain traction~\cite{kirillov2023segment,ma2024segment,pan2023ai}, the need for training methodologies compatible with mixed image modalities is becoming increasingly critical. In such contexts, the data-driven design of S2S2 offers significant benefits, as integrating knowledge from different domains into a single training strategy is challenging. We validate our proposed strategy across popular network architectures and demonstrate its effectiveness in improving both in-domain performance and single-source domain generalization across various CT, MRI, and RGB images. 

Our main contributions are summarized as follows:
\begin{itemize}
\item We propose a versatile add-on training strategy, semantic stacking, that enhances robustness without requiring specialized domain knowledge.
\item We provide theoretical analysis enabling a practical, efficient method for learning the semantic stacking that scales to large datasets.
\item We demonstrate our method's ability to improve both in-domain performance and out-of-domain robustness.
\end{itemize}

\section{Related Work}
\subsection{Data Augmentation in Medical Image Analysis}
Early approaches adapted data augmentation strategies from natural images to medical images~\cite{ronneberger2015u, milletari2016v}. For example, nnU-net~\cite{isensee2021nnu} employed a predefined pipeline with operations like rotation, scaling, Gaussian noise, and blur. Inspired by AutoAug~\cite{cubuk2018autoaugment}, later approaches explored automated data augmentation strategies, but these relied on traditional spatial and color transformations~\cite{xu2020automatic, qin2020automatic, lyu2022aadg, yang2019searching}.

While traditional methods are simple and effective, they fail to fully exploit the distinctive characteristics of medical images. Recently, generative models, such as GANs~\cite{goodfellow2020generative, denton2015deep, beers2018high, yi2019generative} and diffusion models~\cite{ho2020denoising, rombach2022high, kazerouni2022diffusion}, have been used for synthesizing medical images. Most generative approaches have focused on classification tasks~\cite{pinaya2022brain, khader2023denoising, tang2023multi, ye2023synthetic, peng2023generating, deo2023shape}, with segmentation tasks receiving less attention. Notable exceptions include brainSPADE~\cite{fernandez2022can}, which trained segmentation models solely with synthetic data, and DPGAN~\cite{chai2022synthetic}, which used synthetic augmentation to address class imbalance. However, these methods exhibit limitations in performance and applicability. Our work addresses this gap by proposing a versatile add-on training strategy that enhances both in-domain performance and out-of-domain robustness.

\subsection{Single-Source Domain Generalization}
Domain generalization aims to train models that perform reliably on previously unseen data distributions~\cite{wang2022generalizing,zhou2022domain}. Our goal aligns with single-source domain generalization (SDG), where models are trained without access to target or additional source domain information. Recent SDG approaches use specialized augmentations, adapt model architectures, or propose unique training methods~\cite{zhou2022domain,su2023rethinking,xu2020robust,zhou2022generalizable,huang2020self,hu2023devil,guo2024infproto,liao2024dual}. Different from these methods that only focus of out-of-domain robustness, our approach provides a versatile add-on strategy that enhances model robustness and semantic representation without changing existing augmentations, architectures, or training paradigms, ensuring strong in-domain performance while preparing models for deployment across varied medical imaging domains.

\section{Method}
\label{sec:method}
\begin{figure}[ht!]
    \centering
    \includegraphics[width=\linewidth]{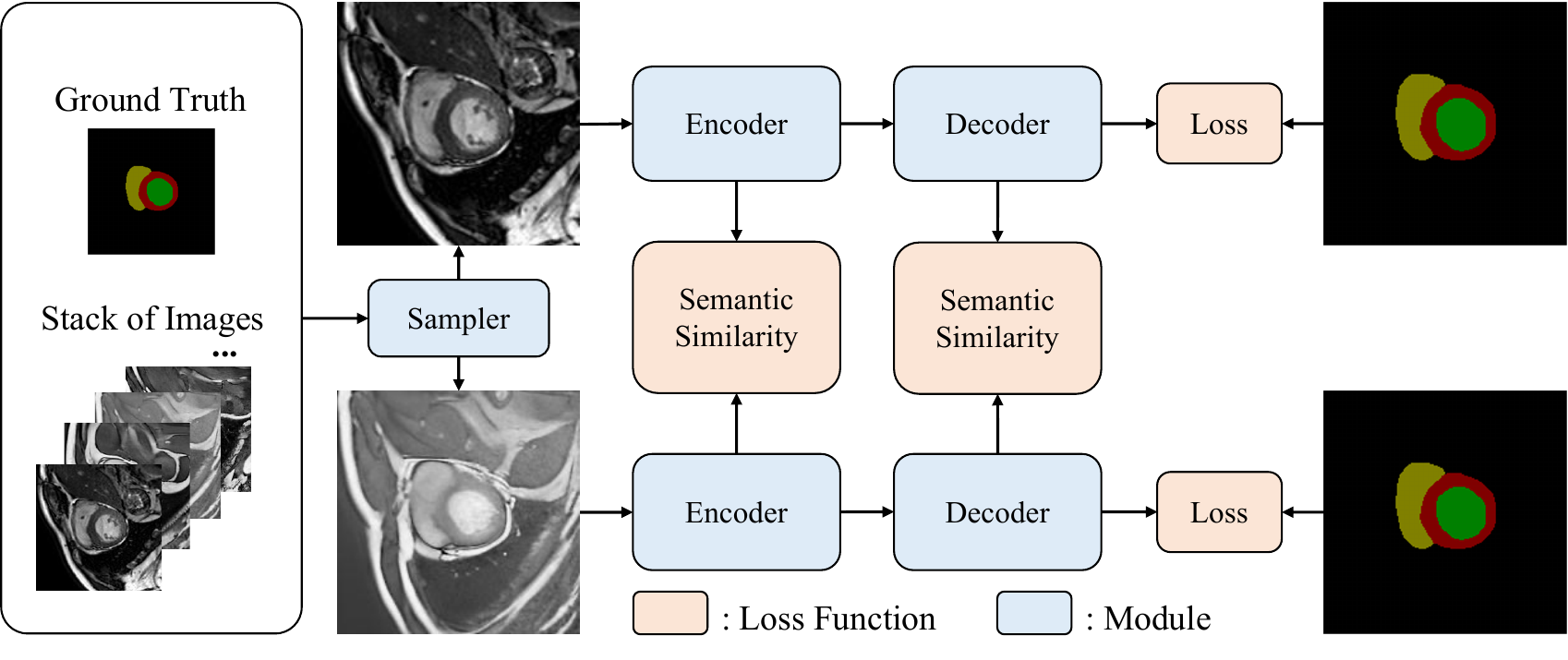}
    \caption{Illustration of the proposed S2S2 framework. A stack of images given is generated from the ground truth semantic segmentation map. Two samples from the stack are then fed into the network, where the training process is guided by the consistency between features alongside the segmentation loss.}
    \label{fig:method}
\end{figure}

\subsection{From Image Stacking to Semantic Stacking}

In semantic segmentation, our objective is to recover the ground truth segmentation map ${y}$ from an input image $x$. This entails classifying each pixel in the semantic feature map $t$ to yield the segmentation map $y=\mathcal{H}(t)$, where $\mathcal{H}$ denotes a classifier. Ideally, the goal is to minimize the discrepancy between the estimated semantic feature map $\hat{t}$ and the truth but unknown $t$. Since $t$ itself is not directly observable, our practical objective shifts to reducing the difference between the estimated segmentation map $\hat{y}$ and the true segmentation map $y$. However, because the classifier $\mathcal{H}$ may map different inputs to the same output label, suggesting that the feature map derived from the training data guided by pixel-level supervision may inherently carry bias. To address this, we leverage the concept of image stacking, a technique traditionally utilized in image denoising, to obtain a more accurate approximation of $t$.

In image denoising, as depicted in Fig.~\ref{fig:semantic_stacking_compare}~(a), the primary objective is to estimate the unknown ground truth image $x$. Image stacking employs multiple noisy images to approximate the ground truth image. Let $\{x_1, \cdots, x_n\}$ represent a collection of noisy images sampled from $\mathcal{N}(x,\sigma_x)$, where $x$ denotes the ground truth image and $\sigma_x$ the noise variance. Let $\hat{x}=\mathcal{P}\left (x_1, \cdots, x_n\right )$ denotes the pooled result of the image stack using mean or median pooling method $\mathcal{P}$, then
\begin{equation}
    \hat{x} \sim \mathcal{N}\left(x,\frac{\sigma_x}{\sqrt{n}}\right)\;.
\end{equation}
As $n$ grows, the precision of the estimated image $\hat{x}$ relative to the ground truth image $x$ enhances.

Adapting this principle for semantic feature estimation, as illustrated in Fig.~\ref{fig:semantic_stacking_compare}~(b), allows us to approach semantic feature mapping with a novel perspective. Specifically, for a given network $\mathcal{F}$ without regularization, we can acquire a semantic feature map $t_i = \mathcal{F}(x_i)\sim \mathcal{N}(t,\sigma)$, where $t$ represents the ground truth semantic feature map. Following the same principle as in image stacking, if we possess a collection of semantic features $\{t_1, \cdots, t_n\}$ corresponding to the identical semantic feature map, pooling these features as $\hat{t}=\mathcal{P}(t_1, \cdots, t_n)$ yields an estimated feature map with diminished variance, expressed as:
\begin{equation}
\label{eqn:stack_mean}
\hat{t} \sim \mathcal{N}\left( t,\frac{\sigma}{\sqrt{n}}\right)\;.
\end{equation}
Let $\mathcal{D}$ denote a distance metric. Utilizing $\hat{t}$ as an approximation of $t$ allows for the optimization of $\mathcal{D}(t_i,\hat{t})$ to enhance the training of $\mathcal{F}$, aiming for $\mathcal{F}$ to generate an accurate $\hat{t}$.

\subsection{Practical Objective for Semantic Stacking}

Direct approximation of $\hat{t}$ from $t$ necessitates constructing a stack of $n$ feature maps, which becomes impractical for large $n$ due to the need for multiple activation copies. To overcome this, we use Bayesian updating. Given a sequence of feature maps $\{t_1, \cdots, t_n\}$, the estimated posterior distribution of $\hat{t}$ is defined as:
\begin{equation}
    \mathbb{E}[\hat{t}]=\frac{\sigma^2{t}_{0}+\sigma_0^2\sum^n_{i=1} t_i}{\sigma^2+n\sigma_0^2}\;, \quad
    \mathrm{Var}[\hat{t}]=\frac{\sigma^2\sigma_0^2}{\sigma^2+n\sigma_0^2}\;,
\end{equation}
where $\sigma_0$ and $t_0$ are the prior distribution's hyperparameters. Assuming $\mathcal{D}$ satisfies the triangle inequality and adopting the $L_1$ distance for simplicity, minimizing $\mathcal{D}(t_i,\mathbb{E}[\hat{t}])$ is achieved through the following optimization:
\begin{gather}
\begin{aligned}
    \mathcal{D}\left(t_i,\mathbb{E}[\hat{t}]\right)
    &=\left | t_i-\frac{\sigma^2{t}_{0}+\sigma_0^2\sum^n_{j=1}t_j}{\sigma^2+n\sigma_0^2} \right |\\
    &\mathllap{=} \frac{1}{\sigma^2+n\sigma_0^2} \left |{\sigma^2(t_i-{t}_{0})+\sigma_0^2\sum^n_{j\ne i}(t_i-t_j)} \right |\\
    &\mathllap{\le} \frac{\sigma^2}{\sigma^2+n\sigma_0^2}|t_i-{t}_{0}|+\frac{\sigma_0^2}{\sigma^2+n\sigma_0^2}\sum^n_{j\ne i}|t_i-t_j|\\
    &\mathllap{\le} \frac{\sigma^2}{\sigma^2+n\sigma_0^2} \mathcal{D}(t_i,{t}_{0})+\frac{\sigma_0^2}{\sigma^2+n\sigma_0^2}\sum^n_{j\ne i} \mathcal{D}(t_i,t_j)\\
\end{aligned}
\end{gather}

We observe that $\mathcal{D}(t_i,\mathbb{E}[\hat{t}])$ is upper-bounded by a weighted sum of all $\mathcal{D}(t_i,t_j)$.  Therefore, minimizing $\mathcal{D}(t_i,\mathbb{E}[\hat{t}])$ effectively requires minimizing $\mathcal{D}(t_i,t_j)$ between any pair of feature maps in the stack. This insight permits sampling just two images at a time from the stack and minimizing the distance between their corresponding feature maps. The resulting semantic consistency loss is formulated as:
\begin{equation}
\mathcal{L}_\text{sc}=\mathcal{D}\left (\mathcal{F}(x_i),\mathcal{F}(x_j) \right )\;,
\end{equation}
where $\mathcal{D}$ is a suitable distance metric that adheres to the triangle inequality, with $x_i,x_j$ being two distinct samples from the stack of images corresponding to the same segmentation map. This methodology, termed S2S2, is illustrated in Fig.~\ref{fig:method}.

\subsection{Constructing Semantic Stack}

Generating images that align with a specific semantic segmentation map poses a significant challenge, particularly in medical image analysis, where annotations are costly and scarce. Recent advances in generative models have provided new ways for synthesizing realistic medical images. In contrast to traditional photometric adjustments like intensity or scale~\cite{cai2023sbss} changes that only account for variations due to equipment differences, variations in human organs are can be learned and simulated using generative models. Utilizing a conditional image generation approach, we generate a set of images based on a given segmentation map. This generative strategy not only enhances diversity but also reduces reliance on dataset-specific knowledge, such as particular intensity variations or color shifts introduced in methods like SLAug~\cite{su2023rethinking}, thereby offering a more generalized solution. Specifically, we fine-tune a Stable Diffusion model~\cite{rombach2022high}, employing ControlNet~\cite{zhang2023adding} for segmentation map control. Although the synthesized images might not precisely replicate the ground truth distribution, we suggest that generating a substantial volume of high-quality images can improve model performance. 

After generating a series of semantic feature maps $\{t_1, \cdots, t_n\}\sim \mathcal{N}( t^g,{\sigma^g})$ from the synthesized images, where $\sigma^g$ reflects the variance indicative of the generated feature maps' quality, and $t^g$ represents the mean, we posit that $t^g \approx t$. This assumption rests on the premise that fine-tuning the generative model with accurate ground truth annotations aligns the mean of the generated feature maps with the ground truth mean, while the variance captures residual discrepancies. In line with previous formulations (Eq.~\ref{eqn:stack_mean}), we have:
$\hat{t}^g \sim \mathcal{N}( t,\frac{\sigma^g}{\sqrt{n}})$.
Specifically, if $\frac{\sigma^g}{\sqrt{n}}\le \sigma$, then $\hat{t}^g$ offers a more accurate estimate of the ground truth semantic feature map, which indicates the potential of enhancing model performance through the minimization of $\mathcal{D}(t_i,\hat{t}^g)$. Although empirically validating this condition may be challenging due to the unknown values of $\sigma^g$ and $\sigma$, theoretical guarantees ensure its validity as $n$ increases.

\section{Dataset}
To comprehensively evaluate the efficacy of our method across diverse medical image segmentation scenarios, we conducted experiments assessing both in-domain and out-of-domain performance. These evaluations covered a variety of imaging modalities, including RGB, CT, and MRI. Details on data prepossessing are in the Appendix.

For RGB images, we utilized two polyp segmentation datasets: CVC-ClinicDB~\cite{cvc} and Kvasir-SEG~\cite{kvasir}. CVC-ClinicDB comprises 612 labeled images, while Kvasir-SEG includes 1,000 labeled images. These datasets, originating from distinct sites and captured using different devices, provide variability in the data. The processing of RGB datasets adhered to the methods described in previous studies~\cite{sanderson2022fcn}. For CT images, we evaluated using the Synapse multi-organ segmentation dataset~\footnote{\url{https://www.synapse.org/\#!Synapse:syn3193805/wiki/217789}}, which includes 30 abdominal CT scans with comprehensive annotations for multi-organ segmentation tasks. In the MRI category, our evaluation encompassed several datasets focused on abdominal and cardiac segmentation. The Combined Healthy Abdominal Organ Segmentation (CHAOS)~\cite{kavur2021chaos} dataset consists of 20 T2-SPIR MRI images focused on abdominal organ segmentation. For cardiac segmentation, we included a dataset~\cite{zhuang2022cardiac} comprising 45 late gadolinium enhanced (LGE) MRI images and 45 balanced steady-state free precession (bSSFP) MRI images, alongside the Automatic Cardiac Diagnosis Challenge (ACDC)~\cite{bernard2018deep} dataset, which features 100 cases of Cine MRI images. 

\section{Results}
Only average metrics are reported in this section for clarity; the class-specific metrics are detailed in the Appendix. Since S2S2 is applicable to any method, we evaluate its performance on representative methods and include baseline methods as references. These baseline methods include MSRF-Net~\cite{srivastava2021msrf} and PraNet~\cite{fan2020pranet} for the Kvasir and CVC datasets; R50-AttnUNet~\cite{schlemper2019attention}, ViT-CUP~\cite{dosovitskiy2020image}, and R50-ViT-CUP~\cite{dosovitskiy2020image} for the Synapse and ACDC datasets; and Cutout~\cite{devries2017improved}, RSC~\cite{huang2020self}, MixStyle~\cite{zhou2021domain}, AdvBias~\cite{carlucci2019domain}, RandConv~\cite{xu2020robust}, and CSDG~\cite{ouyang2022causality} for abdominal and cardiac datasets.

\subsection{Implementation Details}
We compared S2S2 against several established approaches in medical image analysis, as well as a state-of-the-art technique in single-source domain generalization. These established techniques serve as baseline methods for our experiments. All experimental procedures adhered to the methodologies outlined by these baselines, with exceptions made solely for components that integrate our proposed approach (detailed in the Appendix). Synthetic images were generated using Stable Diffusion 2.5 fine-tuned on training images with segmentation-map-controlled ControlNet for 100 epochs. Further details are provided in the Appendix.

Contemporary models for semantic segmentation are typically comprised of an encoder for capturing high-level semantics and a decoder for pixel-level details. We hypothesize that both levels of features are useful and apply our semantic consistency loss to both components, denoted as $\mathcal{L}_\text{sc}^\text{enc}$ and $\mathcal{L}_\text{sc}^\text{dec}$, respectively. The final loss function is formulated as 
\begin{equation}
\mathcal{L}=\mathcal{L}_\text{seg}+\alpha^\text{enc}\mathcal{L}_\text{sc}^\text{enc}+\alpha^\text{dec}\mathcal{L}_\text{sc}^\text{dec}\;,
\end{equation}
where $\mathcal{L}_\text{seg}$ represents the segmentation loss derived from any chosen method. The variables $\alpha^\text{enc}$ and $\alpha^\text{dec}$ are the weights for the consistency losses. For simplicity, we define the distance function as $\mathcal{D}(t_i,t_j)=1-\mathrm{CosSim}(t_i,t_j)$ where $\mathrm{CosSim}$ is cosine similarity.

\subsection{In-domain Performance}
As an add-on method, our foundational premise posits that the integration of S2S2 should not detrimentally affect the performance of the baseline method within the scope of in-domain evaluation. To verify this, we rigorously evaluated S2S2 across a variety of acclaimed network architectures on datasets derived from RGB, CT, and MRI images. Furthermore, we aim to underscore the advantages of adopting a universally applicable method over approaches that are narrowly tailored to specific tasks. To this end, we incorporated SLAug~\cite{su2023rethinking}, a state-of-the-art method devised for enhancing single-domain generalization in CT/MRI imaging, into our in-domain benchmarks.

\begin{table}[!ht]
\centering
\setlength{\tabcolsep}{1mm} 
\centering\small
\begin{tabular}{l|cc|c}
\hline

{Method} & {Synapse} & {ACDC} & Mean\\  
\hline
R50-{AttnUNet} & 75.57 & 86.75 & 81.16\\
ViT-CUP & 67.86 & 81.45 & 74.66\\
R50-ViT-CUP & 71.29 & 87.57 &  79.43\\ \hline
{{TransUNet}} & \underline{76.86}& \underline{88.86} & 82.86\\
{+S2S2} & \textbf{81.19}& \textbf{90.40} & \textbf{85.80}\scalebox{0.75}{$+ 2.94$}\\
\hline
\end{tabular}
\caption{In-domain performance comparison on the Synapse multi-organ CT dataset and ACDC dataset. Dice score (\%) is used as the evaluation metric. The best-performing method is highlighted in bold, and the second-best is underlined. The improvement achieved by S2S2 is indicated.}
\label{tab:transunet_indomain}
\end{table}

As shown in Table~\ref{tab:transunet_indomain}, the integration of S2S2 significantly elevates the in-domain performance for CT/MRI datasets on widely recognized models. Similarly, Table~\ref{tab:polyp_in_domain} demonstrates that the deployment of S2S2 concurrently amplifies the efficacy of FCBFormer on RGB datasets.

Notably, the baseline methods already incorporate augmentation techniques such as color space and spatial augmentation, indicating that S2S2 operates independently of the baseline method or image modality. The semantic stack provides a superior representation of the ground truth semantic feature map than the original unconstrained semantic feature map. We observe an enhanced performance with an increase in the number of classes, potentially attributable to the generative model's refined control over image generation or the amplified complexity of maintaining semantic consistency across broader classes. 

\begin{figure}[!ht]
\centering
    \includegraphics[width=0.65\linewidth]{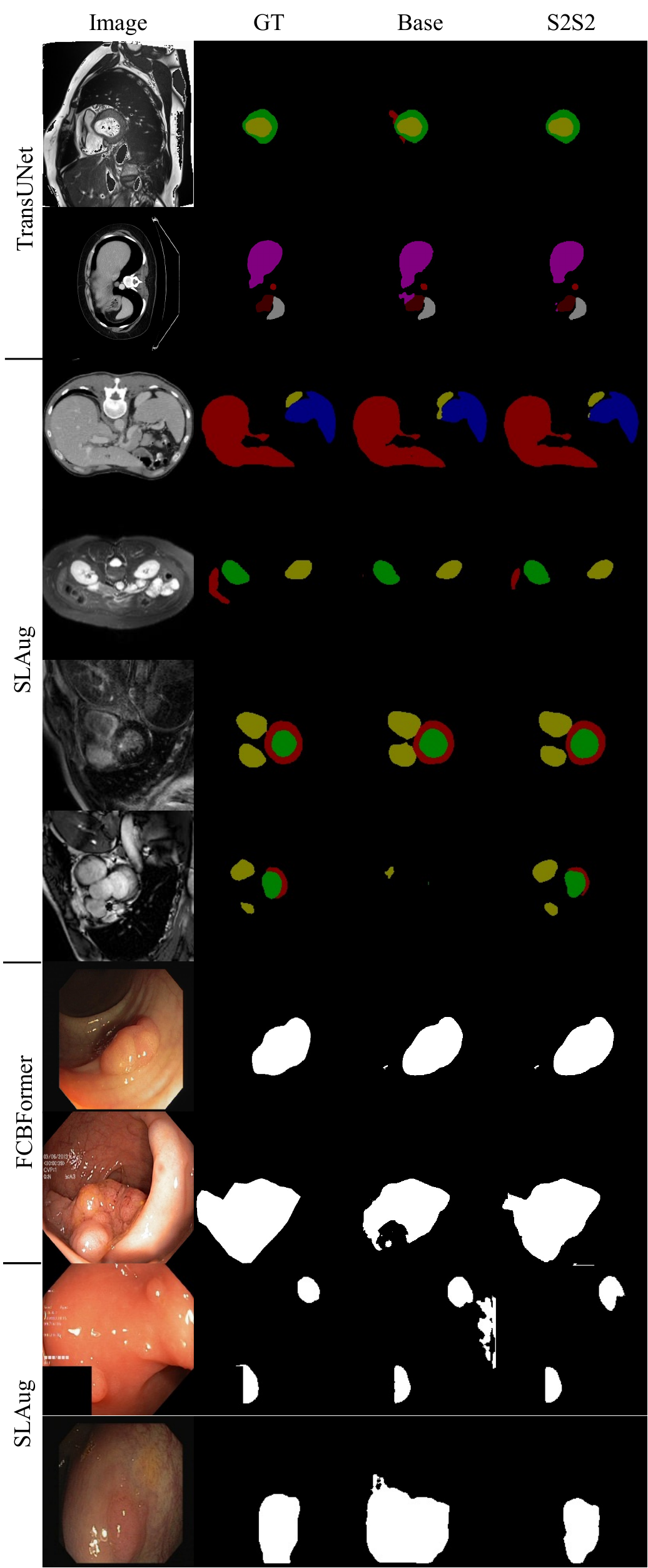}
    \caption{Visualization of the improvement achieved by applying S2S2 to the base method in the in-domain setting. `GT' is the ground truth. `Base' refers to the corresponding method without S2S2.}
    \label{fig:qualitative_in_domain}
\end{figure}

\begin{table}[!ht]
\centering\small
\begin{tabular}{l|cc|c}
\hline
Method & Kvasir & CVC & Mean\\ \hline
MSRF-Net & \underline{92.17} & \underline{94.20} & \underline{93.19}\\ 
PraNet & 89.80 & 89.90 & 89.90\\ \hline
SLAug & 84.85 & 85.39 & 85.12\\ 
SLAug+S2S2 & 85.33 & 88.76 & 87.05\scalebox{0.75}{$+ 1.93$}\\ \hline
FCBFormer & 91.90 & 93.46 & {92.68}\\ 
+S2S2 & \textbf{93.20} & \textbf{94.88} & \textbf{94.04}\scalebox{0.75}{$+ 1.36$}\\ \hline
\end{tabular}
\caption{In-domain performance comparison on RGB datasets. Dice score (\%) is used as the evaluation metric.}
\label{tab:polyp_in_domain}
\end{table}

\begin{table}[!t]
\small
\centering\small
\setlength{\tabcolsep}{1mm} 
\begin{tabular}{l|cc|cc|c}\hline
\multirow{2}{*}{Method} & \multicolumn{2}{c|}{Abdominal} & \multicolumn{2}{c|}{Cardiac} & \multirow{2}{*}{Mean}\\
\cline{2-5}
 & CT & MRI & bSSFP & LGE &\\ \hline
Supervised (CSDG) & \textbf{89.74} & \underline{90.85} & 88.16 & \textbf{88.15} & \textbf{89.23}\\ \hline
SLAug & 82.66 & 90.60 & \textbf{92.27} & 87.35 & 88.22\\
+S2S2 & \underline{84.21} & \textbf{91.28} & \underline{92.16} & \underline{87.62} & \underline{88.82}\scalebox{0.75}{$+ 0.60$}\\ 
\hline
\end{tabular}
\caption{In-domain performance comparison on slices of 3D medical image datasets. Dice score (\%) is used as the evaluation metric.}
\label{tab:in-domain-slaug}
\end{table}

In addition, our analysis reveals that SLAug~\cite{su2023rethinking}, despite being specifically engineered for CT/MRI imaging modalities through the exploitation of domain-specific knowledge, fails to deliver comparable benefits for RGB imaging (Table~\ref{tab:polyp_in_domain}). However, the subsequent application of S2S2 atop SLAug results in a discernible enhancement in performance metrics, indicating that S2S2 introduces an additional layer of supervision beyond the capabilities of domain-specific augmentation techniques. More importantly, even for the CT/MRT images, which SLAug was originally tailored for, S2S2 outperforms the baseline method, as shown in Table~\ref{tab:in-domain-slaug}. This finding suggests that methods focused on domain-specific generalization may inadvertently compromise in-domain performance while optimizing for out-of-domain applicability. In contrast, our approach avoids making assumptions about the application domain, thereby ensuring consistent improvements in in-domain performance across diverse datasets and imaging modalities.

\noindent\textbf{Qualitative Evaluation.} The comparison in Fig.~\ref{fig:qualitative_in_domain} revealed several distinct advantages of our approach. First, S2S2 demonstrates superior capability in identifying the presence or absence of small objects, as evident in rows 1, 4, and 7. Second, it tends to generate smoother segmentation masks, observable in rows 2, 8, and 10. Lastly, S2S2 adopts a more conservative approach in its predictions, particularly highlighted in row 9. 

\subsection{Out-of-domain Performance}
In our out-of-domain evaluations, we benchmarked the S2S2 method against reproducible state-of-the-art, aligning with the settings of FCBFormer~\cite{sanderson2022fcn} for polyp segmentation tasks on RGB images and SLAug~\cite{su2023rethinking} for abdominal organ and cardiac segmentation tasks on CT/MRI images. These comparisons validate not only the robustness of our approach in established domains but also its superior generalization capabilities in unseen domains.

\begin{table}[!ht]
\centering\small
\setlength{\tabcolsep}{1pt} 
\begin{tabular}{l|cc|cc|c}\hline
\multirow{2}{*}{Method} & \multicolumn{2}{c|}{Abdominal} & \multicolumn{2}{c|}{Cardiac} &\multirow{2}{*}{Mean}\\
\cline{2-5}
 & CT-MRI & MRI-CT & bSSFP-LGE & LGE-bSSFP & \\ \hline
Cutout & 80.12 & 70.50 & 78.87 & 85.92  & 78.85\\
RSC & 74.09 & 66.07 & 77.51 & 85.60  & 75.82\\
MixStyle & 77.80 & 63.95 & 75.21 & 86.34  &75.83\\
AdvBias & 80.17 & 64.84 & 79.62 & 86.27  & 77.73\\
RandConv & 80.66 & 76.56 & 83.73 & \underline{87.24}  & 82.05\\
CSDG & {86.31} & {80.40} & {85.01} & 86.99  & 84.68\\ \hline
SLAug & \textbf{88.55} & \underline{81.70} & \textbf{86.42} & 87.17  & \underline{85.96}\\
+S2S2 & \underline{87.75} & \textbf{83.15} & \underline{86.06} & \textbf{87.49}  &\textbf{86.11}\scalebox{0.75}{$+ .15$}\\ 
\hline
\end{tabular}
\caption{Out-of-domain performance comparison on slices of 3D medical image datasets. Dice score (\%) is used as the evaluation metric.}
\label{tab:compare_out_domain}
\end{table}

Beyond demonstrating improvements in in-domain performance, our method also exhibits notable improvements in out-of-domain generalization, as shown in Table~\ref{tab:polyp_generalisability}. Similar to what we observed in in-domain evaluation, the domain-specific method SLAug delivers suboptimal performance on RGB images. However, integrating the proposed S2S2 method fills this gap, enhancing its effectiveness. These results underscore the applicability of S2S2 in augmenting out-of-domain generalization capabilities without necessitating prior insights into the imaging modality or base models. This adaptability renders S2S2 particularly valuable in scenarios where domain-specific knowledge is unavailable. Furthermore, when such expertise is present, domain-specific strategies like SLAug exhibit superior generalization within their intended application domains, as indicated in Table~\ref{tab:compare_out_domain}. While domain-specific approaches are anticipated to excel, the supplementary application of S2S2 on top of SLAug still results in a marginal improvements on both the in-domain and out-of-domain performance. This result consolidates the relevance of S2S2, even in the presence of domain-specific methodologies.

\begin{table}[!t]
\centering\small
\setlength{\tabcolsep}{2pt} 
\begin{tabular}{l|cc|c}
\hline
{Method} & Kvasir-CVC & CVC-Kvasir & Mean\\ \hline
MSRF-Net & 62.38 & 72.96 & 67.67\\ 
PraNet & 79.12 & 79.50 & 79.31\\ \hline
SLAug & 75.62 & 77.09 & 76.36\\ 
+S2S2 & 76.44 & 80.52 & 78.48\scalebox{0.75}{$+ 2.12$}\\ \hline
FCBFormer & \underline{91.16} & \underline{86.46} & \underline{88.81}\\ 
+S2S2 & \textbf{92.85} & \textbf{88.72} & \textbf{90.79}\scalebox{0.75}{$+ 1.98$}\\ 
\hline
\end{tabular}
\caption{Out-of-domain performance comparison on Polyp segmentation (RGB medical image datasets). Dice score (\%) is used as the evaluation metric.}
\label{tab:polyp_generalisability}
\end{table}

\begin{figure}[!ht]
\centering
    \includegraphics[width=0.65\linewidth]{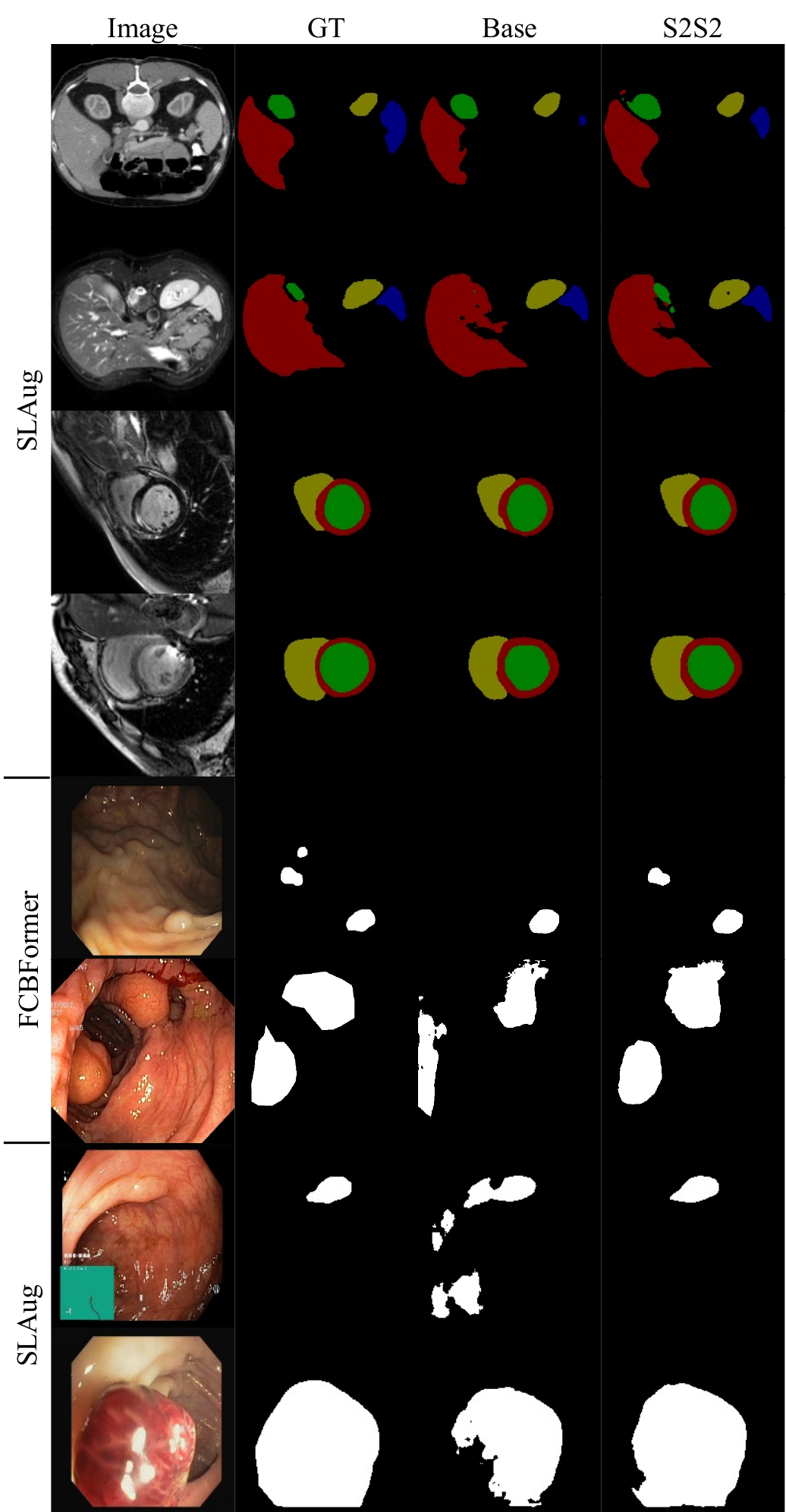}
\caption{Visualization of the improvement achieved by applying S2S2 to the base method in the out-of-domain setting. `GT' is the ground truth. `Base' refers to the corresponding method without S2S2.}
    \label{fig:qualitative_out_of_domain}
\end{figure}

\noindent\textbf{Qualitative Evaluation.} From Fig.~\ref{fig:qualitative_out_of_domain}, we observe similar ability to identify small objects and maintain boundary smoothness in the in-domain samples. Additionally, it is noteworthy that the base method is prone to misclassification issues in RGB images under conditions of significant glare (rows 6 and 8), the presence of unexpected objects (row 7), or insufficient lighting (rows 5 and 8). These conditions introduce what can be considered semantic noise. Our method, designed to mitigate semantic noise within the feature representation, remains robust and unaffected by such artifacts.

\begin{table}[!ht]
\renewcommand\tabcolsep{1mm}
\centering\small
\begin{tabular}{ccc|cc} 
\hline

 Synthetic        & $\mathcal{L}^\text{enc}$    & $\mathcal{L}^\text{dec}$         & ACDC & Synapse     \\
\hline
          &        &        & 88.86 & 76.86   \\
     $\checkmark$     &        &        & 89.66\scalebox{0.75}{$+ .80$} & 77.61\scalebox{0.75}{$+ .75$} \\
     $\checkmark$     &   $\checkmark$     &        & 90.64\scalebox{0.75}{$+ 1.78$}  & 80.29\scalebox{0.75}{$+ 3.43$}\\
     $\checkmark$    &   $\checkmark$   &   $\checkmark$    & 90.40\scalebox{0.75}{$+ 1.54$} &  81.19\scalebox{0.75}{$+ 4.33$} \\
\hline
\end{tabular}
\caption{Performance of TransUNet using different proposed modules, measured in DSC (\%). `Synthetic' indicates the use of synthetic images. $\mathcal{L}^\text{enc}$ denotes the application of consistency loss on encoder features. $\mathcal{L}^\text{dec}$ denotes the application of consistency loss on decoder features.}
\label{tab:transunet_ablation}
\end{table}

\section{Ablation Study}

In our ablation study, we aim to analyze the contribution of each module to performance, as well as the effect of the hyperparameters for the proposed loss. Our strategy to accurately gauge the contributions of our module involves leveraging a baseline model that makes minimal assumptions and favors widespread adoption.
For this purpose, we select TransUNet as the base model, adhering to its established training pipeline. The results of this investigation are detailed in Table~\ref{tab:transunet_ablation}. Employing solely synthetic images in the absence of semantic consistency loss yields a result comparable to the documented in prior works~\cite{pinaya2022brain,khader2023denoising,tang2023multi,ye2023synthetic}, with negligible improvements. The integration of semantic consistency loss $\mathcal{L}^\text{enc}$, however, marks a significant elevation in performance.
Although the subsequent application of $\mathcal{L}^\text{dec}$, in conjunction with $\mathcal{L}^\text{enc}$, results in performance improvement on the Synapse dataset (with 9 classes), a marginal decline in performance is observed on the ACDC dataset (with 4 classes). This result is consistent with our earlier insight, indicating the superiority of the S2S2 method in datasets characterized by a greater number of classes. Moreover, the result suggests that the quality of generated images plays a vital role in the method's effectiveness. $\mathcal{L}^\text{enc}$ performs on a higher level semantic feature that is less sensitive to low-level detail of the generated images whereas $\mathcal{L}^\text{dec}$ operates on the pixel level that is very sensitive to the low-level detail. This dynamic is reflected in our experiments, wherein the inclusion of $\mathcal{L}^\text{dec}$ may potentially detract from out-domain performance. Nonetheless, the application of any form of semantic consistency loss invariably transcends the performance of the baseline model, underscoring the overall efficacy of the proposed S2S2 method.

\begin{figure}[h!]
\centering
    \includegraphics[width=0.9\linewidth]{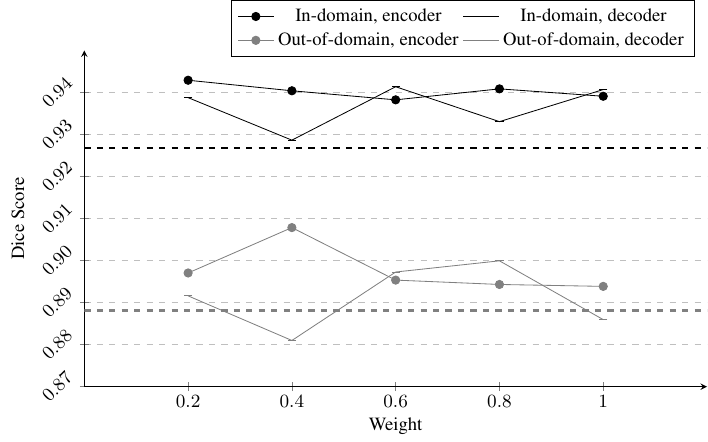}
\caption{Ablation study results using FCBFormer with the proposed S2S2 method. Dashed lines indicate the performance of the base method.}
\label{fig:alphas_ablation}
\end{figure}

To further investigate the impact of loss weighting on performance in both in-domain and out-of-domain contexts, we conducted an ablation study using FCBFormer on RGB images. We measured the Dice score on both in-domain and out-of-domain datasets, focusing on the effects of $\alpha^\text{enc}$ and $\alpha^\text{dec}$. Each variable was analyzed in isolation by setting the alternative to zero for individual assessments. From the analysis presented in Fig.~\ref{fig:alphas_ablation}, it is observed that $\alpha^\text{enc}$ exerts a relatively consistent influence on in-domain performance, with the most notable improvement in out-domain performance is observed at $\alpha^\text{enc}=0.4$. In contrast, the impact of $\alpha^\text{dec}$ appears less consistent, with the greatest fluctuations occurring within the range $\alpha^\text{dec}\in [0.2,0.6]$ for both in-domain and out-of-domain datasets. This discrepancy in the behavior of losses on top of the encoder and decoder may stem from the generative model's capacity to more effectively capture higher-level semantic details as opposed to lower-level information, thereby rendering the encoder features more stable than those of the decoder, which aligns with our previous results. Moreover, the decoder features are subjected to additional layers of network weights, potentially amplifying errors inherent within the network architecture. This result suggests a preference for $\mathcal{L}^\text{enc}$ over $\mathcal{L}^\text{dec}$, attributed to its reduced sensitivity to variations in image quality. Despite the distinct behaviors observed, both semantic consistency losses contribute to the overall enhancement in model performance. Finally, if we apply the semantic consistency loss with only photometric augmentation such as Gaussian blur and color jitters, we get worse performance than the base method (detailed in the Appendix). This result further suggests the importance of the semantic stacking in addition to traditional augmentation.

\section{Discussion and Conclusion}
We introduce S2S2, a novel and broadly applicable add-on training strategy inspired by the image stacking technique, designed to improve both in-domain performance and out-of-domain robustness. However, the practical application of S2S2 encounters certain constraints. Primarily, the method's reliance on a fine-tuned generative model for semantic stacking, while innovative, introduces computational demands that may limit its suitability for situations with abundant data, such as natural image segmentation tasks. Additionally, the performance of S2S2 is inherently tied to the generative model's effectiveness across various datasets, which could significantly influence outcomes.

In conclusion, our findings present a compelling case for S2S2 as a powerful complement to existing domain-specific augmentation methods and architectural modifications. This strategy not only enhances model robustness but also represents a meaningful step toward the development of universally applicable solutions in image segmentation.

\section*{Acknowledgements}
Research reported in this publication was supported by the National Institute of Biomedical Imaging and Bioengineering of the National Institutes of Health (NIH) under award R01EB030130. The content is solely the responsibility of the authors and does not necessarily represent the official views of the NIH. 
This work used cluster computers at the National Center for Supercomputing Applications through an allocation from the Advanced Cyberinfrastructure Coordination Ecosystem: Services \& Support (ACCESS) program, which is supported by National Science Foundation (NSF) grants 2138259, 2138286, 2138307, 2137603, and 2138296. The work also used the Extreme Science and Engineering Discovery Environment (XSEDE) under NSF grant 1548562.

\clearpage
\bibliography{aaai25}

\appendix
\section{Dataset Details}
\subsection{Class Definition and Visualization Palette}
\begin{figure*}[ht!]
\centering
   \includegraphics[width=0.9\linewidth]{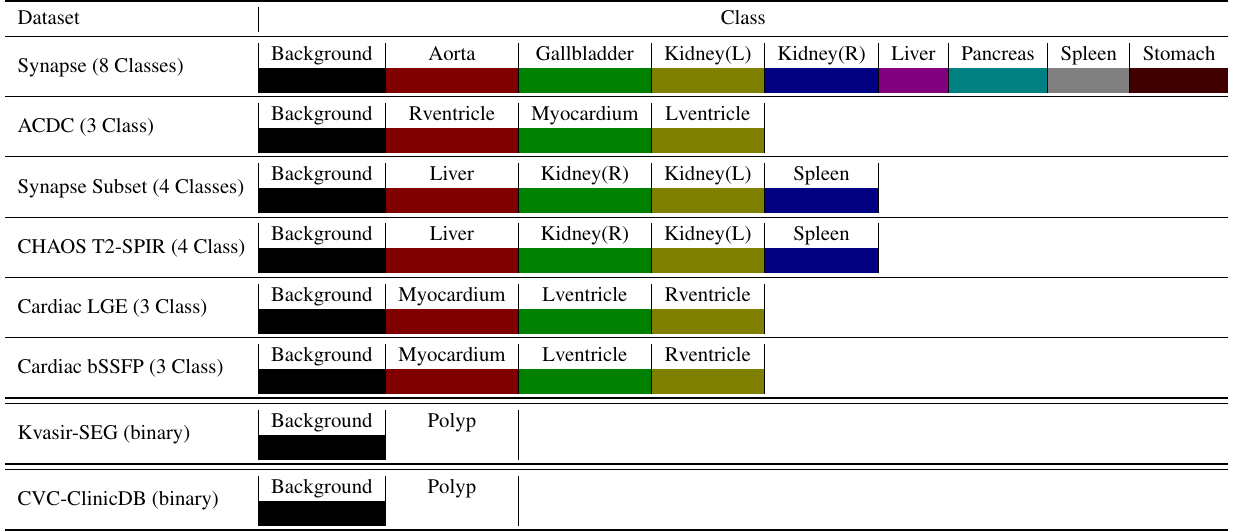}
    \caption{Unique color assignments for classes in medical image segmentation datasets.}
    \label{tab:color_pallete}
\end{figure*}

To facilitate a comprehensive understanding and consistent visualization of segmentation results across different datasets, we assign a unique color to each class within our datasets, as illustrated in Fig.~\ref{tab:color_pallete}. The classes, along with their corresponding abbreviations where applicable, include: Right Ventricle (RVC), Myocardium (MYO), Left Ventricle (LVC), Aorta, Gallbladder, Kidney (Left), Kidney (Right), Liver, Pancreas, Spleen, Stomach, and Polyp.

\subsection{Pre-processing}
For comprehensive and standardized evaluation, we adhere to specific pre-processing protocols across different datasets. The procedures for Synapse (8 classes) and the ACDC datasets follow the guidelines established in~\cite{chen2021transunet}. Specifically, for the Synapse dataset, we employ a random split of 18 training cases (comprising 2,212 axial slices) and 12 validation cases. For the ACDC dataset, the division consists of 70 training cases (1,930 axial slices), 10 validation cases, and 20 test cases. Image values are constrained within the range [-125, 275], and each 3D image is normalized to the range [0, 1]. We apply both spatial and color space augmentations to these datasets.

For Synapse (4 classes), CHAOS T2-SPIR, LGE, and bSSFP datasets, our pre-processing aligns with the methods described in~\cite{ouyang2022causality}. In the Synapse (4 classes) dataset, we implement a windowing technique with Housefield values set between [-125, 275]. For the CHAOS T2-SPIR, LGE, and bSSFP datasets, the top 0.5\% of the histogram values are clipped, each 3D image is normalized to have zero mean and unit variance, and similar to other datasets, both spatial and color space augmentations are utilized.

The Kvasir-SEG and CVC-ClinicDB datasets undergo pre-processing as per the protocols in~\cite{sanderson2022fcn}, with all RGB images normalized to the range of [-1, 1] and subjected to both spatial and color space augmentations.

\section{Metrics}
To quantify the performance of our segmentation models, we utilize a set of standardized metrics across our experiments. These include Dice score (Dice), intersection over union (IoU), precision (Prec), recall (Rec), and Hausdorff distance in millimeters (HD). For 3D images (Synapse, CHAOS T2-SPIR, LGE, and bSSFP datasets), these metrics are calculated over the entire 3D volume, whereas for 2D images in the Kvasir-SEG and CVC-ClinicDB datasets, evaluations are performed on individual images. Consistent with~\cite{sanderson2022fcn}, we employ the prefix ``m'' (e.g., mDice) to denote the mean scores for metrics in the polyp segmentation datasets.

\section{Implementation Details}
\subsection{Semantic Stack Generation}
We leverage Stable Diffusion (SD) 2.1, fine-tuned specifically to our training datasets, to generate synthetic images. This process is augmented with segmentation-map-controlled ControlNet, enabling precise adherence to the ground truth segmentation maps during synthetic image generation. The resizing and fine-tuning parameters are carefully chosen based on dataset characteristics and prior literature.

For datasets including Synapse (8 classes), ACDC, Kvasir-SEG, and CVC-ClinicDB, we standardize the image dimensions to \(512 \times 512\), aligning with the native resolution of SD 2.1. For Synapse (4 classes), CHAOS T2-SPIR, LGE, and bSSFP, the images are adjusted to \(192 \times 192\), as suggested by~\cite{su2023rethinking,ouyang2022causality}. ControlNet is fine-tuned over 100 epochs with a batch size of 16 and a learning rate of \(1e-5\), with the SD parameters frozen to ensure consistency. Distinct models are trained for each dataset to mitigate the risk of test domain data leakage.

The control mechanism for synthetic image generation leverages the ground truth segmentation maps, coupled with structured text descriptions detailed in Table~\ref{tab:gen_prompt}, as prompts. This methodological choice is aimed at enhancing the relevance and accuracy of the generated images.

\begin{figure*}[p]
    \centering
    \includegraphics[width=0.95\linewidth]{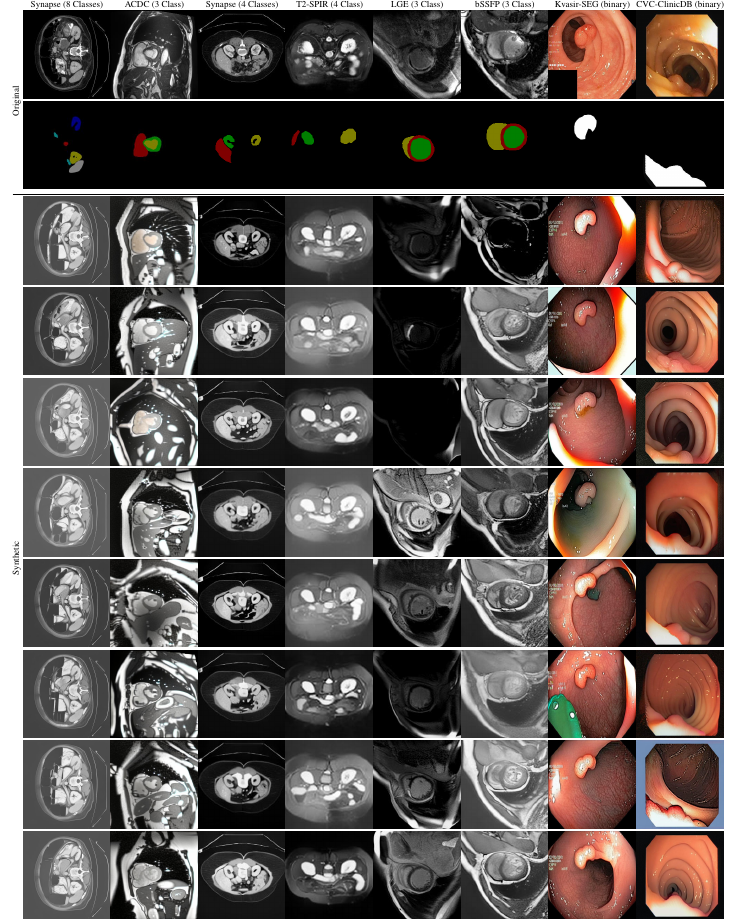}
    \caption{Visualization of synthetic images generated.}
    \label{fig:gen_images}
\end{figure*}

For constructing semantic stacks, we opt for a stack size of \(n=16\) synthetic images for each ground truth segmentation mask within the training set. The sampling process utilizes a denoising diffusion implicit model, executed over 50 steps with a strength setting of 1.0, scale of 9.0, and eta of 0.0. To ensure experimental repeatability, the random seed is maintained consistently throughout all experiments. Examples of the generated images are presented in Fig.~\ref{fig:gen_images}, illustrating the efficacy and precision of the synthetic image generation process.

The training pipeline illustrated in Fig.~\ref{fig:s2s2_alg}. The simplicity allows for versatile application.
\begin{figure}[ht!]
    \centering
\lstset{
    language=Python,
    basicstyle=\ttfamily\tiny,
    keywordstyle=\color{blue},
    stringstyle=\color{red},
    commentstyle=\color{green!50!black},
    mathescape=true,
    numbers=left,
    numberstyle=\tiny,
    stepnumber=1,
    frame=lines
}
\begin{lstlisting}
    """
        Only need two images for each mask 
        at each iteration base on Sec 3.2.
        We use the original image as one 
        of the images
    """
    for image_0, mask in dataset:
        image_1 = finetuned_gen_model(mask)
        # encode the images
        enc_feat_0 = seg_encoder(image_0)
        enc_feat_1 = seg_encoder(image_1)
        # decode the encoder features
        dec_feat_0 = seg_decoder(enc_feat_0)
        dec_feat_1 = seg_decoder(enc_feat_1)
        # pixel-level classification
        logits_0 = linear(dec_feat_0)
        logits_1 = linear(dec_feat_1)
        # compute the segmentation loss
        loss = seg_loss(image_0,mask) + seg_loss(image_1,mask)
        # compute the encoder consistency loss
        loss += alpha_enc * enc_consist(enc_feat_0,enc_feat_1)
        # compute the decoder consistency loss
        loss += alpha_dec * enc_consist(dec_feat_0,dec_feat_1)
        # update the model parameters
        loss.backward()
        optimizer.step()
\end{lstlisting}
    \caption{Pseudocode for S2S2 training.}
    \label{fig:s2s2_alg}
\end{figure}

\begin{table*}[ht!]
    \centering\small
    \setlength{\tabcolsep}{1pt} 
    \begin{tabular}{l}
    \toprule
         Dataset \\
         \midrule
         {Synapse~\footnote{\url{https://www.synapse.org/\#!Synapse:syn3193805/wiki/217789}} (8 Classes)} \\
         A 2D slice of an abdomen CT scan showing [class names].\\
         \midrule
        {ACDC~\cite{bernard2018deep} (3 Class)} \\
        A 2D slice of a cardiac MRI scan showing [class names].\\
         \midrule
         {Synapse (4 Classes)} \\
         A 2D slice of an abdominal CT scan showing [class names].\\
         \midrule
         {T2-SPIR~\cite{kavur2021chaos} (4 Class)} \\
         A 2D slice of an abdominal T2-SPIR MRI scan showing [class names].\\
         \midrule
         {LGE~\cite{zhuang2022cardiac} (3 Class)} \\
         A 2D slice of a cardiac MRI scan using balanced steady-state free precession showing [class names]. \\
         \midrule
         {bSSFP~\cite{zhuang2022cardiac} (3 Class)} \\
         A 2D slice of a cardiac MRI scan using late gadolinium enhanced showing [class names]. \\
         \midrule
         {Kvasir-SEG~\cite{kvasir} (binary)} \\
         An image of the human gastrointestinal tract captured by colonoscope showing [class names].\\
         \midrule
         {CVC-ClinicDB~\cite{cvc} (binary)} \\
         An image of the human gastrointestinal tract captured by colonoscope [class names].\\
         \bottomrule
    \end{tabular}
    
    \caption{Prompts used for the generative model for each dataset. Top row: the dataset and class name. Bottom row: the corresponding text prompt. The final prompt is created by concatenating the dataset-specific prompt with the class names of the class names present in the image.}
    \label{tab:gen_prompt}
\end{table*}

\subsection{Evaluation Methodology}
Our evaluation strategy strictly adheres to the foundational training and testing parameters established by the respective base methods. This section delineates only the distinctions introduced by the implementation of our S2S2 strategy. In all experiments, alongside the original image, we uniformly select a single image from the generated stack for analysis.

For TransUNet~\cite{chen2021transunet}, experiments are standardized with $\alpha^\text{enc}=1$ and/or $\alpha^\text{dec}=1$ to maintain simplicity in variable adjustment. When incorporating FCBFormer~\cite{sanderson2022fcn}, which features dual encoders, we extend the application of the semantic similarity loss across both encoders and the decoder for comprehensive ablation studies. In the final evaluation, the loss is specifically applied to the decoder with $\alpha^\text{dec}=0$ and two encoders with $\alpha^\text{enc}_1=0.4$ and $\alpha^\text{enc}_2=0.4$, optimizing for balanced performance enhancement.

For experiments using SLAug~\cite{su2023rethinking}, we consistently apply $\alpha^\text{enc}=0.1$ and $\alpha^\text{dec}=0$ across all trials. Notably, our methodology demonstrated a tendency for achieving heightened in-domain performance relatively early in the training cycle, likely attributable to an increased initial loss magnitude. Consequently, we opt for an early stopping of the training process at 1,100 epochs for our method, as opposed to extending to the full 2,000 epochs. However, applying early stopping to the SLAug baseline negatively affects performance. Therefore, for SLAug, we adhere to the original epoch settings to preserve the integrity of comparative analysis. To apply SLAug on RGB images, we train the model for 500 epochs on all the experiments.

\section{Additional Results}

\begin{table*}[!ht]
\centering\small
\setlength{\tabcolsep}{1pt}
\begin{tabular}{l|cccccccc|cc}
\hline
\multirow{2}{*}{Method} & \multirow{2}{*}{Aorta} & \multirow{2}{*}{Gallbladder} & \multirow{2}{*}{L-Kidney} & \multirow{2}{*}{R-Kidney} & \multirow{2}{*}{Liver} & \multirow{2}{*}{Pancreas} & \multirow{2}{*}{Spleen} & \multirow{2}{*}{Stomach} & \multicolumn{2}{c}{{Average}} \\
 \cline{10-11} 
 &          &         &             &            &             &           &             &           & Dice~$\uparrow$ & HD~$\downarrow$ \\ 
\hline
R50-{AttnUNet}~\cite{schlemper2019attention} & 55.92      & 63.91          & 79.20         & 72.71          & 93.56       & 49.37        & 87.19       & 74.95        & 75.57   & 36.97        \\
ViT~\cite{dosovitskiy2020image}            & 44.38      & 39.59          & 67.46         & 62.94          & 89.21       & 43.14       & 75.45       & 69.78        & 61.50        & 39.61        \\
ViT-CUP~\cite{dosovitskiy2020image}              & 70.19      & 45.10          & 74.70         & 67.40         & 91.32       & 42.00        & 81.75       & 70.44       & 67.86        & 36.11        \\
R50-ViT-CUP~\cite{dosovitskiy2020image}              & 73.73      & 55.13          & 75.80         & 72.20         & 91.51       & 45.99        & 81.99      & 73.95        & 71.29        & 32.87        \\
{{TransUNet}}~\cite{chen2021transunet} & {86.81}       & {56.82}          & {81.99}          & {78.13}         & {93.95}      & {55.44}         & {85.07}      & {76.64}        & \underline{76.86}        & \underline{26.73}        \\ 
{{TransUNet}+S2S2} & {87.52}        & {63.40}          & {86.39}          & {82.61}          & {94.76}    & {64.55}         & {89.41}      & {80.87}        & \textbf{81.19}        & \textbf{24.81}        \\ 
\hline
\end{tabular}
\caption{In-domain performance comparison on the Synapse multi-organ CT dataset across baseline architectures. The average Dice score (\%), average Hausdorff distance (mm), and Dice score (\%) for each organ are reported. The best-performing method is highlighted in bold, and the second-best is underlined.}
\label{tab:synapse}
\end{table*}

\begin{table*}[!ht]
\centering\small
\setlength{\tabcolsep}{1pt}
\begin{tabular}{l|ccc|c} 
\hline

Method         & RVC           & MYO        & LVC   & Average          \\
\hline
R50-AttnUNet~\cite{schlemper2019attention}      & 87.58       & 79.20       & 93.47   & 86.75       \\
ViT-CUP~\cite{dosovitskiy2020image}       & 81.46	& 70.71	 & 92.18 & 81.45    \\
R50-ViT-CUP~\cite{dosovitskiy2020image}       &  86.07	& {81.88}	& 94.75 & 87.57 \\ \hline
TransUNet~\cite{chen2021transunet}         &  {89.28}       & {81.80}       & {95.49} &  \underline{88.86}    \\
TransUNet+S2S2         & {88.95}       & {86.16}       &  {96.07} & \textbf{90.40}    \\
\hline
\end{tabular}
\caption{In-domain performance comparison on the ACDC dataset in Dice score (\%). The best-performing method is highlighted in bold, and the second-best is underlined.}
\label{tab:acdc}
\end{table*}

\begin{table*}[!ht]
\centering\small
\setlength{\tabcolsep}{1pt}
\begin{tabular}{l|ccccc|cccc}\hline
\multirow{2}{*}{Method} & \multicolumn{5}{c|}{Abdominal CT (Synapse)} & \multicolumn{4}{c}{Cardiac bSSFP} \\
\cline{2-10}
 & Liver & R-Kidney & L-Kidney & \multicolumn{1}{c|}{Spleen} & Average & LVC & MYO & \multicolumn{1}{c|}{RVC} & Average \\ \hline
Supervised~\cite{ouyang2022causality} & {98.87} & {92.11} & {91.75} & {88.55} & \textbf{89.74} & 91.16 & 82.93 & \multicolumn{1}{c|}{90.39} & 88.16 \\ 
SLAug~\cite{su2023rethinking} & 96.48 & 66.97 & 79.24 & \multicolumn{1}{c|}{87.92} & 82.66 & {95.55} & {88.10} & \multicolumn{1}{c|}{{93.14}} & \textbf{92.27} \\
SLAug+S2S2& {96.60} & {67.66} & {83.58} & \multicolumn{1}{c|}{{89.01}} & \underline{84.21} & {95.85} & {87.58} & \multicolumn{1}{c|}{{93.05}} & \underline{92.16} \\ \hline\hline
\multirow{2}{*}{Method} & \multicolumn{5}{c|}{Abdominal MRI (T2-SPIR)} & \multicolumn{4}{c}{Cardiac LGE} \\
\cline{2-10}
 & Liver & R-Kidney & L-Kidney & \multicolumn{1}{c|}{Spleen} & Average & LVC & MYO & \multicolumn{1}{c|}{RVC} & Average \\ \hline
Supervised~\cite{ouyang2022causality} & 91.30 & {92.43} & {89.86} & \multicolumn{1}{c|}{{89.83}} & {90.85} & 92.04 & 83.11 & \multicolumn{1}{c|}{89.30} & \textbf{88.15} \\ 
SLAug~\cite{su2023rethinking} & {91.75} & {92.29} & {91.14} & \multicolumn{1}{c|}{{87.22}} & {90.60} & 89.31 &81.50 & \multicolumn{1}{c|}{91.25} & 87.35\\
SLAug+S2S2 & {91.69} & {91.84} & {90.72} & \multicolumn{1}{c|}{{90.88}} & \textbf{91.28} & {89.55} & {81.83} & \multicolumn{1}{c|}{{91.47}} &\underline{87.62}\\  \hline
\end{tabular}
\caption{In-domain performance comparison on slices of 3D medical image datasets. Dice score (\%) is used as the evaluation metric. The best-performing method is highlighted in bold, and the second-best is underlined.}
\label{tab:in-domain-slaug_app}
\end{table*}

\begin{table*}[!ht]
\centering\small
\setlength{\tabcolsep}{1pt}
\begin{tabular}{l|cccc|cccc}
\hline
\multirow{2}{*}{Method}                                     & \multicolumn{4}{c|}{Kvasir-SEG \cite{kvasir}} & \multicolumn{4}{c}{CVC-ClinicDB \cite{cvc}} \\ \cline{2-9}
                                      & Dice         & IoU        & Prec.        & Rec.        & Dice        & IoU        & Prec.        & Rec.        \\ \hline
MSRF-Net \cite{srivastava2021msrf}    & 92.17             & \textbf{89.14}           & \textbf{96.66}                & 91.98             & 94.20            & \textbf{90.43}           & \underline{94.27}                & \textbf{95.67}             \\
PraNet \cite{fan2020pranet}                 & 89.80 & 84.00                & -             & - & 89.90 & 84.90 & -  & -           \\ 
\hline
SLAug~\cite{su2023rethinking}                     & 84.85 & 77.3 & 88.12 & 84.75            & 85.39         & 76.98           & 82.43                & 91.37             \\
SLAug+S2S2                          & 85.33           & 78.00           & 86.58  &  86.91            & 88.76             & 81.01           & 88.73               & 90.34            \\ 
FCBFormer~\cite{sanderson2022fcn}                     & 91.90             & 87.05           & 94.05               &   91.62           & 93.46             & 89.17           & 93.57               &   93.66  \\
FCBFormer+S2S2                         & \textbf{93.20}             & \underline{88.57}         & \underline{94.54}               & \textbf{93.59}             & \textbf{94.88}           & \underline{90.41}           & \textbf{94.63}                &\underline{95.43}             \\ \hline
\end{tabular}
\caption{In-domain performance comparison on RGB datasets. Baseline model results are taken from~\cite{sanderson2022fcn}.  Metrics are reported in percentages (\%). The best-performing method is highlighted in bold, and the second-best is underlined.}\label{tab:polyp_in_domain_app}
\end{table*}

\begin{table*}[!ht]
\centering\small
\setlength{\tabcolsep}{1pt}
\begin{tabular}{l|ccccc|cccc}\hline
\multirow{2}{*}{Method} & \multicolumn{5}{c|}{Abdominal CT-MRI} & \multicolumn{4}{c}{Cardiac bSSFP-LGE} \\
\cline{2-10}
 & Liver & R-Kidney & L-Kidney & \multicolumn{1}{c|}{Spleen} & Average & LVC & MYO & \multicolumn{1}{c|}{RVC} & Average \\ \hline
Cutout~\cite{devries2017improved} & 79.80 & 82.32 & 82.14 & \multicolumn{1}{c|}{76.24} & 80.12 & 88.35 & 69.06 & \multicolumn{1}{c|}{79.19} & 78.87 \\
RSC~\cite{huang2020self} & 76.40 & 75.79 & 76.60 & \multicolumn{1}{c|}{67.56} & 74.09 & 87.06 & 69.77 & \multicolumn{1}{c|}{75.69} & 77.51 \\
MixStyle~\cite{zhou2021domain} & 77.63 & 78.41 & 78.03 & \multicolumn{1}{c|}{77.12} & 77.80 & 85.78 & 64.23 & \multicolumn{1}{c|}{75.61} & 75.21 \\
AdvBias~\cite{carlucci2019domain} & 78.54 & 81.70 & 80.69 & \multicolumn{1}{c|}{79.73} & 80.17 & 88.23 & 70.29 & \multicolumn{1}{c|}{80.32} & 79.62 \\
RandConv~\cite{xu2020robust} & 73.63 & 79.69 & 85.89 & \multicolumn{1}{c|}{83.43} & 80.66 & 89.88 & 75.60 & \multicolumn{1}{c|}{85.70} & 83.73 \\
CSDG~\cite{ouyang2022causality} & {86.62} & {87.48} & {86.88} & \multicolumn{1}{c|}{{84.27}} & {86.31} & {90.35} & {77.82} & \multicolumn{1}{c|}{{86.87}} & {85.01} \\\hline
SLAug~\cite{su2023rethinking} & 89.97 & 89.39 & 87.40 & \multicolumn{1}{c|}{87.45} & \textbf{88.55} & 91.56 & 80.28 & \multicolumn{1}{c|}{87.43} & \textbf{ 86.42} \\
SLAug+S2S2& 90.71 & 89.22 & 86.55 & \multicolumn{1}{c|}{84.51} & \underline{87.75} & 91.48 & 79.84 & \multicolumn{1}{c|}{86.87} & \underline{86.06} \\ \hline\hline
\multirow{2}{*}{Method} & \multicolumn{5}{c|}{Abdominal MRI-CT} & \multicolumn{4}{c}{Cardiac LGE-bSSFP} \\
\cline{2-10}
 & Liver & R-Kidney & L-Kidney & \multicolumn{1}{c|}{Spleen} & Average & LVC & MYO & \multicolumn{1}{c|}{RVC} & Average \\ \hline
Cutout~\cite{devries2017improved} & 86.99 & 63.66 & 73.74 & \multicolumn{1}{c|}{57.60} & 70.50 & 90.88 &79.14  & \multicolumn{1}{c|}{87.74} & 85.92 \\
RSC~\cite{huang2020self} & {88.10} & 46.60 & 75.94 & \multicolumn{1}{c|}{53.61} &66.07  & 90.21 & 78.63 & \multicolumn{1}{c|}{87.96} &85.60  \\
MixStyle~\cite{zhou2021domain} & 86.66 & 48.26 & 65.20 & \multicolumn{1}{c|}{55.68} & 63.95 &  91.22&79.64  & \multicolumn{1}{c|}{88.16} & 86.34 \\
AdvBias~\cite{carlucci2019domain} & 87.63 & 52.48 & 68.28 & \multicolumn{1}{c|}{50.95} & 64.84 &91.20  &  79.50& \multicolumn{1}{c|}{88.10} &86.27  \\
RandConv~\cite{xu2020robust} & 84.14 & 76.81 & 77.99 & \multicolumn{1}{c|}{67.32} & 76.56 & {91.98} & {80.92} & \multicolumn{1}{c|}{88.83} &{ \underline{87.24} }\\
CSDG~\cite{ouyang2022causality} & 85.62 & {80.02} & {80.42} & \multicolumn{1}{c|}{{75.56}} & {80.40} &91.37  & 80.43 & \multicolumn{1}{c|}{{89.16}} & 86.99 \\ \hline
SLAug~\cite{su2023rethinking} & {88.87} & {80.23} & {81.59} & \multicolumn{1}{c|}{{76.12}} & \underline{81.70} & 91.43 &80.64 & \multicolumn{1}{c|}{89.43} & 87.17\\
SLAug+S2S2 & 88.30 & 81.79 & 80.31 & \multicolumn{1}{c|}{82.21} & \textbf{83.15} & 92.17 &80.19 & \multicolumn{1}{c|}{90.10} &\textbf{87.49}\\  \hline
\end{tabular}
\caption{Out-of-domain performance on slices of 3D medical image datasets. Dice score (\%) is used as the evaluation metric. The best-performing method is highlighted in bold, and the second-best is underlined.}
\label{tab:compare_out_domain_app}
\end{table*}

\begin{table*}[!ht]
\centering\small
\setlength{\tabcolsep}{1pt}
\begin{tabular}{l|cccc|cccc}
\hline
\multirow{2}{*}{Method}                                  & \multicolumn{4}{c|}{Kvasir-CVC} & \multicolumn{4}{c}{CVC-Kvasir } \\ \cline{2-9}
                                     & Dice         & IoU        & Prec.        & Rec.        & Dice        & IoU        & Prec.        & Rec.        \\ \hline
MSRF-Net \cite{srivastava2021msrf}    & 62.38           & 54.19                & 66.21     & 70.51 & 72.96 & 64.15 & 81.62 & 74.21             \\
PraNet \cite{fan2020pranet}    & 79.12 & 71.19 & 81.52 & 83.16 & 79.50 & 70.73 & 76.87 & \textbf{90.50}             \\
SLAug~\cite{su2023rethinking}                          & 75.62             & 66.97           & 83.19           & 76.65            & 77.09            & 67.91           & 74.34                & \underline{89.11}             \\
SLAug+S2S2                           & 76.44             & 67.81           & 81.97                & 79.37             & 80.52            & 72.14          & 85.49               & 82.12            \\ 
FCBFormer~\cite{sanderson2022fcn}                          & \underline{91.16}             & \underline{85.40}           & \underline{91.89}                & \underline{91.31}             & \underline{86.46}            & \underline{80.27}           & \textbf{92.92}                & 85.22             \\
FCBFormer+S2S2                            & \textbf{92.85}             & \textbf{86.94}           & \textbf{93.46}                & \textbf{92.95}             & \textbf{88.72}            & \textbf{82.79}           & \underline{92.33}                & {88.91}             \\ \hline
\end{tabular}
\caption{Out-of-domain performance on Polyp segmentation (RGB medical image datasets). Metrics are reported in percentages (\%). The best-performing method is highlighted in bold, and the second-best is underlined.}\label{tab:polyp_generalisability_app}
\end{table*}

\begin{table*}[!ht]
\centering\small
\setlength{\tabcolsep}{1pt}
\begin{tabular}{l|cccc|cccc}
\hline
\multirow{2}{*}{Method} & \multicolumn{4}{c|}{CVC Train} & \multicolumn{4}{c}{Kvasir Train} \\ \cline{2-9}
 & Dice & IoU & Precision & Recall & Dice & IoU & Precision & Recall \\ \hline
Baseline (CVC Test)& \textbf{93.46} & \textbf{89.17} & \textbf{93.57} & \textbf{93.66} & \textbf{91.90} & \textbf{87.05} & \textbf{94.05} & \textbf{91.62} \\
\textbf{Aug} (CVC Test) & 84.10 & 79.79 & 85.99 & 85.21 & 85.27 & 79.15 & 93.67 & 82.21 \\ \hline
Baseline (Kvasir Test)& \textbf{91.16} & \textbf{85.40} & \textbf{91.89} & \textbf{91.31} & 91.90 & 87.05 & \textbf{94.05} & \textbf{93.66} \\
\textbf{Aug} (Kvasir Test) & 81.87 & 76.16 & 82.19 & 83.54 & \textbf{92.42} & \textbf{87.82} & 92.73 & \textbf{93.66} \\ \hline
\end{tabular}
\caption{Performance metrics on when applying the semantic consistency loss with only photometric augmentation (Aug). Metrics are reported in percentages (\%). The best-performing method is highlighted in bold.}\label{tab:performance_metrics_app}
\end{table*}

This section provides additional detailed results. All the metrics and classes used in the original work are reported.

Overall, our method improves both in-domain and out-of-domain performance when the base method lacks domain knowledge (e.g., TransUNet, FCBFormer) or is based on incorrect domain assumptions (e.g., SLAug on RGB). Even in scenarios where domain knowledge is available (e.g., SLAug on CT and MRI), our method achieves an average improvement in both in-domain and out-of-domain performance.

However, as shown in Table~\ref{tab:compare_out_domain_app}, our method does not outperform the base method in the CT-MRI and bSSFP-LGE settings. A possible explanation is that the assumptions made by SLAug align better with these specific settings. SLAug incorporates domain knowledge, such as intensity differences between source and target domains, into its augmentation strategy. If these augmentations effectively captures the variation in the target domain, the target domain performance will improve. For instance, SLAug+CT (or SLAug+bSSFP) may better address the variations introduced by MRI (or LGE) than the reverse setup.

As a data-driven method that does not rely on domain knowledge, our approach is not tailored to a specific target domain but is instead designed to enhance robustness across all target domains. Therefore, if the domain-specific augmentation introduced by SLAug already captures the variation in the target domain, the additional application of our method may not provide additional benefits.

\end{document}